\documentclass[preprints,article,accept,moreauthors,pdftex]{Definitions/mdpi}

\firstpage{1} 
\pubvolume{xx}
\issuenum{1}
\articlenumber{5}
\pubyear{2019}
\copyrightyear{2019}
\history{}

\pdfoutput=1

\usepackage{listings}
\usepackage{color}
\usepackage{textcomp}
\definecolor{listinggray}{gray}{0.9}
\definecolor{lbcolor}{rgb}{0.9,0.9,0.9}
\Title{fastai: A Layered API for Deep Learning}

\Author{Jeremy Howard\orcidA{} $^{1,2,\dagger}$ and Sylvain Gugger\orcidB{} $^{1,\dagger}$}
\AuthorNames{Jeremy Howard and Sylvain Gugger}

\address{%
$^{1}$ \quad fast.ai; j@fast.ai, s@fast.ai\\
$^{2}$ \quad University of San Francisco}
\firstnote{These authors contributed equally to this work.}

\abstract{fastai is a deep learning library which provides practitioners with high-level components that can quickly and easily provide state-of-the-art results in standard deep learning domains, and provides researchers with low-level components that can be mixed and matched to build new approaches. It aims to do both things without substantial compromises in ease of use, flexibility, or performance. This is possible thanks to a carefully layered architecture, which expresses common underlying patterns of many deep learning and data processing techniques in terms of decoupled abstractions. These abstractions can be expressed concisely and clearly by leveraging the dynamism of the underlying Python language and the flexibility of the PyTorch library. fastai includes: a new type dispatch system for Python along with a semantic type hierarchy for tensors; a GPU-optimized computer vision library which can be extended in pure Python; an optimizer which refactors out the common functionality of modern optimizers into two basic pieces, allowing optimization algorithms to be implemented in 4-5 lines of code; a novel 2-way callback system that can access any part of the data, model, or optimizer and change it at any point during training; a new data block API; and much more. We have used this library to successfully create a complete deep learning course, which we were able to write more quickly than using previous approaches, and the code was more clear. The library is already in wide use in research, industry, and teaching.}


\begin{document}

\section{Introduction}

fastai is a modern deep learning library, \href{https://github.com/fastai/fastai}{available from GitHub} as open source under the Apache 2 license, which can be installed directly using the conda or pip package managers. It includes \href{https://docs.fast.ai/}{complete documentation and tutorials}, and is the subject of the book \textit{Deep Learning for Coders with fastai and PyTorch: AI Applications Without a PhD}~\cite{dlbook}.

fastai is organized around two main design goals: to be approachable and rapidly productive, while also being deeply hackable and configurable. Other libraries have tended to force a choice between conciseness and speed of development, or flexibility and expressivity, but not both. We wanted to get the clarity and development speed of Keras~\cite{keras} and the customizability of PyTorch. This goal of getting the best of both worlds has motivated the design of a layered architecture. A high-level API powers ready-to-use functions to train models in various applications, offering customizable models with sensible defaults. It is built on top of a hierarchy of lower-level APIs which provide composable building blocks. This way, a user wanting to rewrite part of the high-level API or add particular behavior to suit their needs doesn't have to learn how to use the lowest level.

\begin{figure}[H]
\centering
\includegraphics[width=10cm]{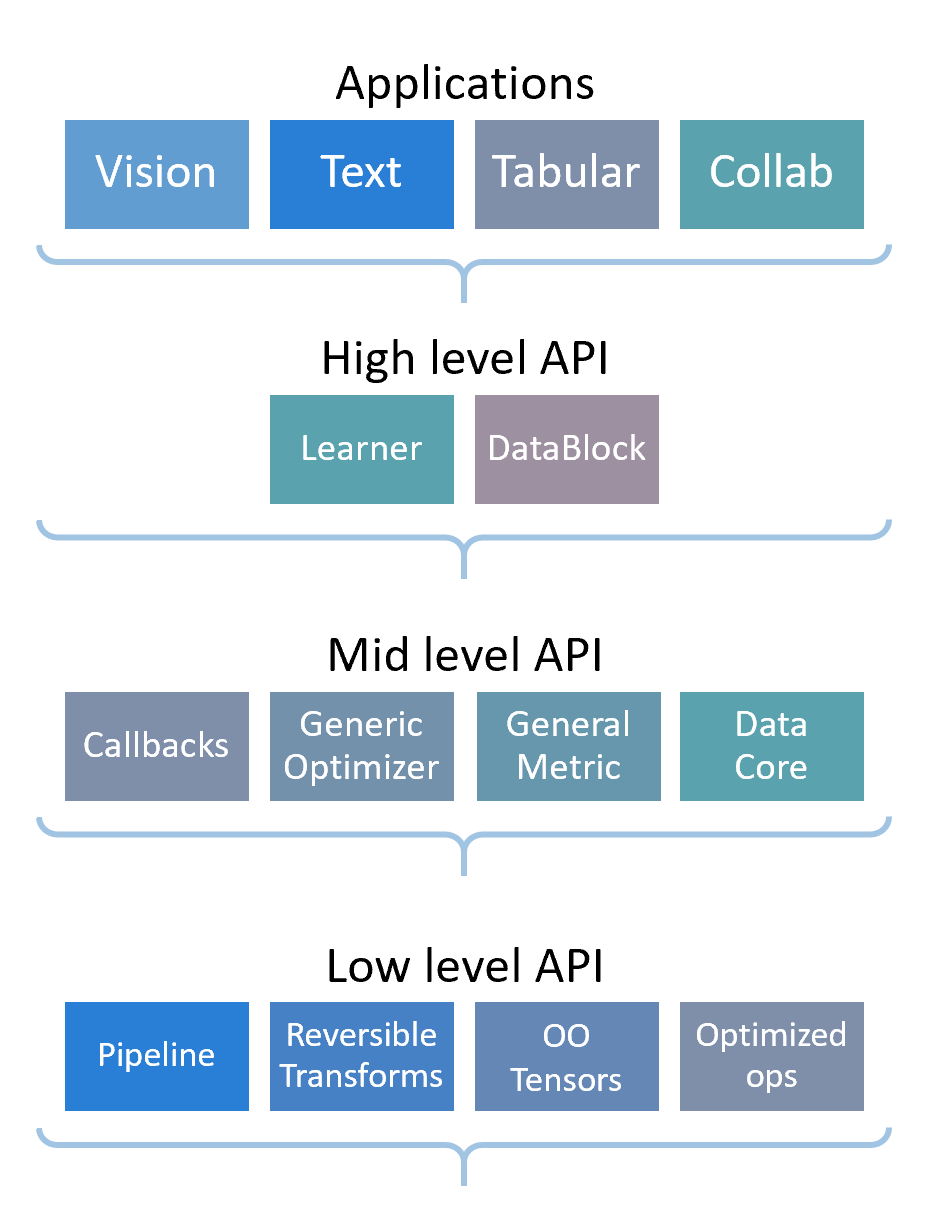}
\caption{The layered API from fastai}
\end{figure}   

The high-level of the API is most likely to be useful to beginners and to practitioners who are mainly in interested in applying pre-existing deep learning methods. It offers concise APIs over four main application areas: vision, text, tabular and time-series analysis, and collaborative filtering. These APIs choose intelligent default values and behaviors based on all available information. For instance, fastai provides a single \verb+Learner+ class which brings together architecture, optimizer, and data, and automatically chooses an appropriate loss function where possible. Integrating these concerns into a single class enables fastai to curate appropriate default choices. To give another example, generally a training set should be shuffled, and a validation does not. So fastai provides a single \verb+DataLoaders+ class which automatically constructs validation and training data loaders with these details already handled. This helps practitioners ensure that they don't make mistakes such as failing to include a validation set. In addition, because the training set and validation set are integrated into a single class, fastai is able, by default, always to display metrics during training using the validation set. 

This use of intelligent defaults--based on our own experience or best practices--extends to incorporating state-of-the-art research wherever possible. For instance, transfer learning is critically important for training models quickly, accurately, and cheaply, but the details matter a great deal. fastai automatically provides transfer learning optimised batch-normalization~\cite{batchnorm} training, layer freezing, and discriminative learning rates~\cite{ulmfit}. In general, the library's use of integrated defaults means it requires fewer lines of code from the user to re-specify information or merely to connect components. As a result, every line of user code tends to be more likely to be meaningful, and easier to read. 

The mid-level API provides the core deep learning and data-processing methods for each of these applications, and low-level APIs provide a library of optimized primitives and functional and object-oriented foundations, which allows the mid-level to be developed and customised.  The library itself is built on top of PyTorch~\cite{pytorch}, NumPy~\cite{numpy}, PIL~\cite{pil}, pandas~\cite{pandas}, and various other libraries. In order to achieve its goal of hackability, the library does not aim to supplant or hide these lower levels or this foundation. Within a fastai model, one can interact directly with the underlying PyTorch primitives; and within a PyTorch model, one can incrementally adopt components from the fastai library as conveniences rather than as an integrated package.

We believe fastai meets its design goals. A user can create and train a state-of-the-art vision model using transfer learning with four understandable lines of code. Perhaps more tellingly, we have been able to implement recent deep learning research papers with just a couple of hours work, whilst matching the performance shown in the papers. We have also used the library for our winning entry in the DawnBench competition~\cite{dawnbench}, training a ResNet-50 on ImageNet to accuracy in 18 minutes.

The following sections describe the main functionality of the various API levels in more detail and review prior related work. We chose to include a lot of code to illustrate the concepts we are presenting. While that code made change slightly as the library or its dependencies evolve (it is running against fastai v2.0.0), the ideas behind stay the same. The next section reviews the high-level APIs "out-of-the-box" applications for some of the most used deep learning domains. The applications provided are vision, text, tabular, and collaborative filtering. 

\section{Applications}

\subsection{Vision}\label{vision}

Here is an example of how to fine-tune an ImageNet~\cite{imagenet} model on the Oxford IIT Pets dataset~\cite{pets} and achieve close to state-of-the-art accuracy within a couple of minutes of training on a single GPU: 
 
\begin{lstlisting}[language=Python]
from fastai.vision.all import *
path = untar_data(URLs.PETS)
dls = ImageDataLoaders.from_name_re(path=path, bs=64,
    fnames = get_image_files(path/"images"), pat = r'/([^/]+)_\d+.jpg$',
    item_tfms=RandomResizedCrop(450, min_scale=0.75), 
    batch_tfms=[*aug_transforms(size=224, max_warp=0.), Normalize.from_stats(*imagenet_stats)])
learn = cnn_learner(dls, resnet34, metrics=error_rate)
learn.fit_one_cycle(4)
\end{lstlisting}

This is not an excerpt; these are all of the lines of code necessary for this task. Each line of code does one important task, allowing the user to focus on what they need to do, rather than minor details. Let's look at each line in turn: 

\begin{lstlisting}[language=Python]
from fastai.vision.all import *
\end{lstlisting}

This first line imports all the necessary pieces from the library. fastai is designed to be usable in a read–eval–print loop (REPL) environment as well as in a complex software system. Even if  using the "import *" syntax is not generally recommended, REPL programmers generally prefer the symbols they need to be directly available to them, which is why fastai supports the "import $*$" style. The library is carefully designed to ensure that importing in this way only imports the symbols that are actually likely to be useful to the user and avoids cluttering the namespace or shadowing important symbols.  

\begin{lstlisting}[language=Python]
path = untar_data(URLs.PETS)
\end{lstlisting}

The second line downloads a standard dataset from the \href{https://course.fast.ai/datasets}{fast.ai datasets collection} (if not previously downloaded) to a configurable location (\verb+~/.fastai/data+ by default), extracts it (if not previously extracted), and returns a \verb+pathlib.Path+ object with the extracted location.

\begin{lstlisting}[language=Python]
dls = ImageDataLoaders.from_name_re(...)
\end{lstlisting}

This line sets up the \verb+DataLoaders+ object. This is an abstraction that represents a combination of training and validation data and will be described more in a later section. \verb+DataLoaders+ can be flexibly defined using the data block API (see \ref{datablock}), or, as here, can be built for specific predefined applications using specific subclasses. In this case, the \verb+ImageDataLoaders+ subclass is created using a regular expression labeller. Many other labellers are provided, particularly focused on labelling based on different kinds of file and folder name patterns, which are very common across a wide range of datasets.

One interesting feature of this API, which is also shared by lower-level fastai data APIs, is the separation of item level and batch level transforms. \emph{Item transforms} are applied, in this case, to individual images on the CPU. \emph{Batch transforms}, on the other hand, are applied to a mini-batch, on the GPU if available. While fastai supports data augmentation on the GPU, images need to be of the same size before being batched. \verb+aug_transforms()+ selects a set of data augmentations that work well across a variety of vision datasets and problems and can be fully customized by providing parameters to the function. This is a good example of a simple "helper function"; it is not strictly necessary, because the user can list all the augmentations that they require using the individual data augmentation classes. However, by providing a single function which curates best practices and makes the most common types of customization available through a single function, users have fewer pieces to learn in order to get good results.

After defining a \verb+DataLoaders+ object the user can easily look at the data with a single line of code: 

\begin{lstlisting}[language=Python]
dls.show_batch()
\end{lstlisting}

\begin{figure}[H]
\centering
\includegraphics[width=7 cm]{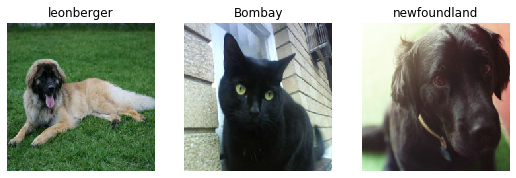}
\caption{A DataLoaders object built with the fastai library knows how to show its elements in a meaningful way. Here the result on the Oxford IIT Pets image classification dataset.}
\end{figure}   

\begin{lstlisting}[language=Python]
learn = cnn_learner(dls, resnet34, metrics=error_rate)
\end{lstlisting}

This fourth line creates a \verb+Learner+, which provides an abstraction combining an optimizer, a model, and the data to train it -- this will be described in more detail in \ref{learner}. Each application has a customized function that creates a \verb+Learner+, which will automatically handle whatever details it can for the user. For instance, in this case it will download an ImageNet-pretrained model, if not already available, remove the classification head of the model, replace it with a head appropriate for this particular dataset, and set appropriate defaults for the optimizer, weight decay, learning rate, and so forth (except where overridden by the user).

\begin{lstlisting}[language=Python]
learn.fit_one_cycle(4)
\end{lstlisting}

The fifth line fits the model. In this case, it is using the 1cycle policy~\cite{onecycle}, which is a recent best practice for training and is not widely available in most deep learning libraries by default. It is annealing both the learning rates, and the momentums, printing metrics on the validation set, displaying results in an HTML table (if run in a Jupyter Notebook, or a console table otherwise), recording losses and metrics after every batch to allow plotting later, and so forth. A GPU will be used if one is available. 

After training a model the user can view the results in various ways, including analysing the errors with \verb+show_results()+:

\begin{lstlisting}[language=Python]
learn.show_results()
\end{lstlisting}

\begin{figure}[H]
\centering
\includegraphics[width=7 cm]{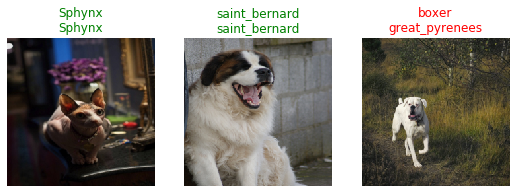}
\caption{A Learner knows from the data and the model type how to represent the results. It can even highlight model errors (here predicted class at bottom and actual at top).}
\end{figure}   

Here is another example of a vision application, this time for segmentation on the CamVid dataset~\cite{camvid}: 

\begin{lstlisting}[language=Python]
from fastai.vision.all import *
path = untar_data(URLs.CAMVID)
dls = SegmentationDataLoaders.from_label_func(path=path, bs=8,
    fnames = get_image_files(path/"images"), 
    label_func = lambda o: path/'labels'/f'{o.stem}_P{o.suffix}',
    codes = np.loadtxt(path/'codes.txt', dtype=str),                         
    batch_tfms=[*aug_transforms(size=(360,480)), Normalize.from_stats(*imagenet_stats)])
learn = unet_learner(dls, resnet34, metrics=acc_segment)
learn.fit_one_cycle(8, pct_start=0.9)
\end{lstlisting}
 
The lines of code to create and train this model are almost identical to those for a classification model, except for those necessary to tell fastai about the differences in the processing of the input data. The exact same line of code that was used for the image classification example can also be used to display the segmentation data:

\begin{figure}[H]
\centering
\includegraphics[width=7 cm]{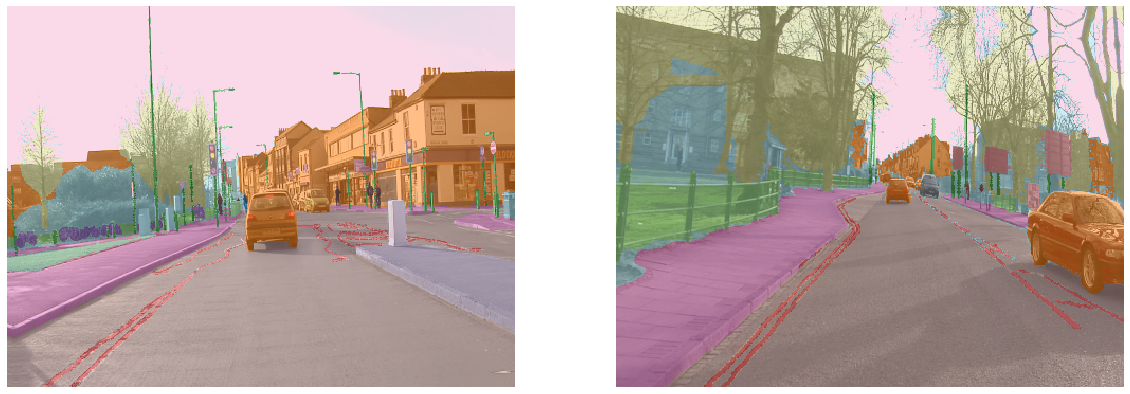}
\caption{In this case, fastai knows that the data is for a segmentation task, and therefore it color-codes and overlays, with transparency, the segmentation layer on top of the input images.}
\end{figure}   

Furthermore, the user can also view the results of the model, which again are visualized automatically in a way suitable for this task: 

\begin{figure}[H]
\centering
\includegraphics[width=7 cm]{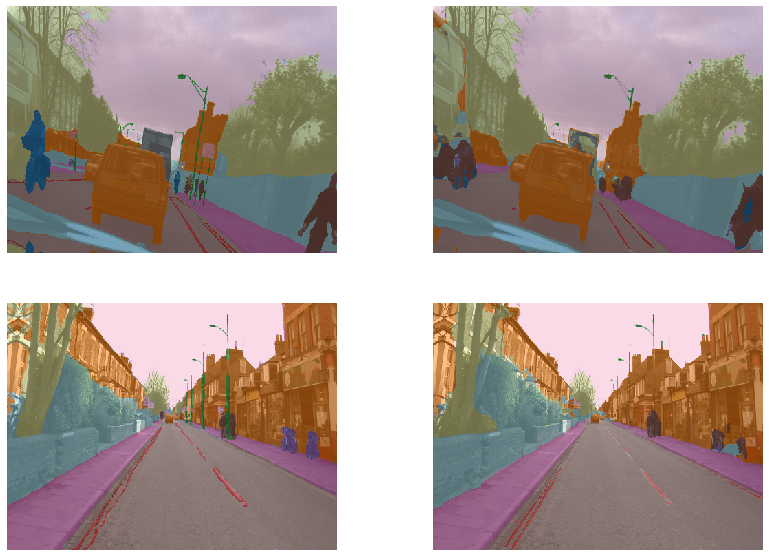}
\caption{For a segmentation task, the ground-truth mask is laid at the right of the predicted mask.}
\end{figure}   

\subsection{Text}\label{text}

In modern natural language processing (NLP), perhaps the most important approach to building models is through fine-tuning pre-trained language models. To train a language model in fastai requires very similar code to the previous examples (here on the IMDb dataset~\cite{imdb}): 

\begin{lstlisting}[language=Python]
from fastai.text.all import *
path = untar_data(URLs.IMDB_SAMPLE)
df_tok,count = tokenize_df(pd.read_csv(path/'texts.csv'), ['text'])
dls_lm = TextDataLoaders.from_df(df_tok, path=path,
    vocab=make_vocab(count), text_col='text', is_lm=True)
learn = language_model_learner(dls_lm, AWD_LSTM, metrics=Perplexity()])
learn.fit_one_cycle(1, 2e-2, moms=(0.8,0.7,0.8))
\end{lstlisting}

Fine-tuning this model for classification requires the same basic steps: 

\begin{lstlisting}[language=Python]
dls_clas = TextDataLoaders.from_df(df_tok, path=path
    vocab=make_vocab(count), text_col='text', label_col='label')
learn = text_classifier_learner(dls_clas, AWD_LSTM, metrics=accuracy)
learn.fit_one_cycle(1, 2e-2, moms=(0.8,0.7,0.8))
\end{lstlisting}

The same API is also used to view the \verb+DataLoaders+: 

\begin{figure}[H]
\centering
\includegraphics[width=15 cm]{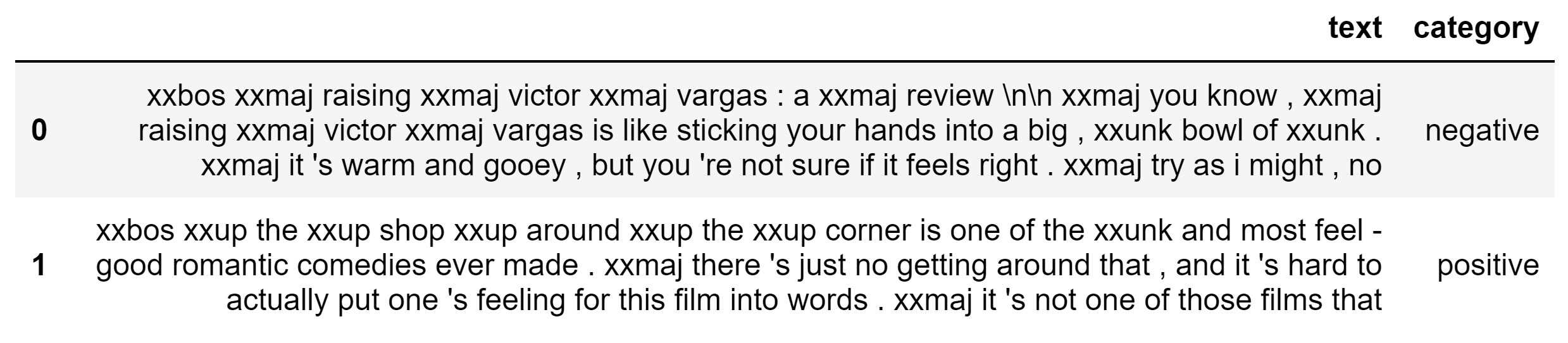}
\caption{In text classification, the batches are shown in a DataFrame with the tokenized texts.}
\end{figure}  

The biggest challenge with creating text applications is often the processing of the input data. fastai provides a flexible processing pipeline with predefined rules for best practices, such as handling capitalization by adding tokens. For instance, there is a compromise between lower-casing all text and losing information, versus keeping the original capitalisation and ending up with too many tokens in your vocabulary. fastai handles this by adding a special single token representing that the next symbol should be treated as uppercase or sentence case and then converts the text itself to lowercase. fastai uses a number of these special tokens. Another example is that a sequence of more than three repeated characters is replaced with a special repetition token, along with a number of repetitions and then the repeated character. These rules largely replicate the approaches discussed in~\cite{ulmfit} and are not normally made available as defaults in most NLP modelling libraries. 

The tokenization is flexible and can support many different organizers. The default used is Spacy. A SentencePiece tokenizer~\cite{sentencepiece} is also provided by the library. Subword tokenization~\cite{WuSCLNMKCGMKSJL16}~\cite{abs-1804-10959}, such as that provided by SentencePiece, has been used in many recent NLP breakthroughs~\cite{radford2019language}~\cite{abs-1810-04805}. 

Numericalization and vocabulary creation often requires many lines of code, and careful management here fails and caching. In fastai that is handled transparently and automatically. Input data can be provided in many different forms, including: a separate file on disk for each document, delimited files in various formats, and so forth. The API also allows for complete customisation. SentencePiece is particularly useful for handling multiple languages and was used in MultiFIT~\cite{multifit}, along with fastai, for this purpose. This provided models and state-of-the-art results across many different languages using a single code base. 

fastai's text models are based on AWD-LSTM~\cite{awdlstm}. The user community have provided external connectors to the popular HuggingFace Transformers library~\cite{transformers}. The training of the models proceeds in the same way as for the vision examples with defaults appropriate for these models automatically selected. We are not aware of other libraries that provide direct support for transfer learning best practices in NLP, such as those shown in~\cite{ulmfit}. Because the tokenisation is built on top of a layered architecture, users can replace the base tokeniser with their own choices and will automatically get support for the underlying parallel process model provided by fastai. It will also automatically handle serialization of intermediate outputs so that they can be reused in future processing pipelines. 

The results of training the model can be visualised with the same API as used for image models, shown in a way appropriate for NLP: 

\begin{figure}[H]
\centering
\includegraphics[width=15 cm]{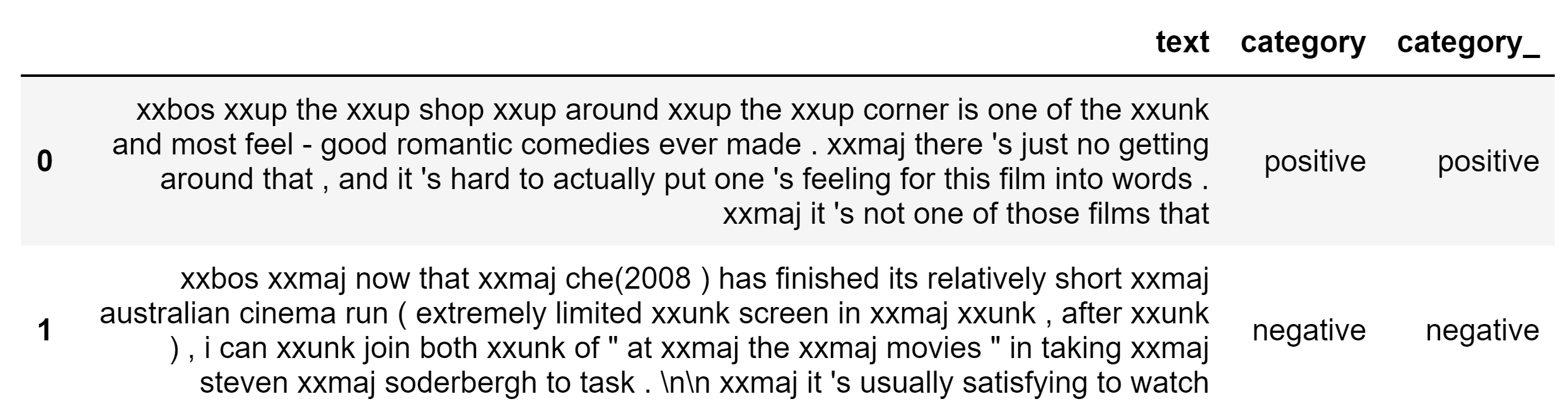}
\caption{In text classification, results are displayed in a DataFrame with the tokenized texts.}
\end{figure} 

\subsection{Tabular}\label{tabular}

Tabular models have not been very widely used in deep learning; Gradient boosting machines and similar methods are more commonly used in industry and research settings. However, there have been examples of competition winning approaches and academic state-of-the-art results using deep learning~\cite{BrebissonSAVB15}. Deep learning models are particularly useful for datasets with high cardinality categorical variables because they provide embeddings that can be used even for non-deep learning models~\cite{GuoB16}. One of the challenges is there has not been examples of libraries which directly support best practices for tabular modelling using deep learning. 

The pandas library~\cite{pandas} already provides excellent support for processing tabular data sets, and fastai does not attempt to replace it. Instead, it adds additional functionality to pandas DataFrames through various pre-processing functions, such as automatically adding features that are useful for modelling with date data. fastai also provides features for automatically creating appropriate \verb+DataLoaders+ with separated validation and training sets, using a variety of mechanisms, such as randomly splitting rows, or selecting rows based on some column. 

The code to create and train a model suitable for this data should look familiar, there is just information specific to tabular data requires when building the \verb+DataLoaders+ object.

\begin{lstlisting}[language=Python]
from fastai2.tabular.all import *
path = untar_data(URLs.ADULT_SAMPLE)
df = pd.read_csv(path/'adult.csv')
dls = TabularDataLoaders.from_df(df, path, 
    procs=[Categorify, FillMissing, Normalize],
    cat_names=['workclass', 'education', 'marital-status', 'occupation', 'relationship', 'race'], 
    cont_names=['age', 'fnlwgt', 'education-num'],
    y_names='salary', valid_idx=list(range(1024,1260)), bs=64)
learn = tabular_learner(dls, layers=[200,100], metrics=accuracy)
learn.fit_one_cycle(3)
\end{lstlisting}

As for every other application, \verb+dls.show_batch+ and \verb+learn.show_results+ will display a DataFrame with some samples.

fastai also integrates with NVIDIA's cuDF library, providing end-to-end GPU optimized data processing and model training. fastai is the first deep learning framework to integrate with cuDF in this way. 

\subsection{Collaborative filtering}\label{collab}

Collaborative filtering is normally modelled using a probabilistic matrix factorisation approach~\cite{mnih2008probabilistic}. In practice however, a dataset generally has much more than just (for example) a user ID and a product ID, but instead has many characteristics of the user, product, time period, and so forth. It is quite standard to use those to train a model, therefore, fastai attempts to close the gap between collaborative filtering and tabular modelling. A collaborative filtering model in fastai can be simply seen as a tabular model with high cardinality categorical variables. A classic matrix factorisation model is also provided. Both are trained using the same steps that we've seen in the other applications, as in this example using the popular Movielens dataset~\cite{movielens}:

\begin{lstlisting}[language=Python]
from fastai2.collab import *
ratings = pd.read_csv(untar_data(URLs.ML_SAMPLE)/'ratings.csv')
dls = CollabDataLoaders.from_df(ratings, bs=64, seed=42)
learn = collab_learner(dls, n_factors=50, y_range=[0, 5.5])
learn.fit_one_cycle(3)
\end{lstlisting}

\subsection{Deployment}

fastai is mostly focused on model training, but once this is done you can easily export the PyTorch model to serve it in production. The command \verb+Learner.export+ will serialize the model as well as the input pipeline (just the transforms, not the training data) to be able to apply the same to new data.

The library provides \verb+Learner.predict+ and \verb+Learner.get_preds+ to evaluate the model on an item or a new inference DataLoader. Such a DataLoader can easily be built from a set of items with the command \verb+test_dl+.

\section{High-level API design considerations}

\subsection{High-level API foundations}\label{datablock}

The \emph{high-level API} is that which is used by people using these applications. All the fastai applications share some basic components. One such component is the visualisation API, which uses a small number of methods, the main ones being \verb+show_batch+ (for showing input data) and \verb+show_results+ (for showing model results). Different types of model and datasets are able to use this consistent API because of fastai's type dispatch system, a lower-level component which will be discussed in \ref{typedispatch}. The transfer learning capability shared across the applications relies on PyTorch's parameter groups, and fastai's mid-level API then leverages these groups, such as the generic optimizer (see \ref{optimizer}).

In all those applications, the \verb+Learner+ obtained gets the same functionality for the model training. The recommended way of training models using a variant of the 1cycle policy~\cite{onecycle} which uses a warm-up and annealing for the learning rate while doing the opposite with the momentum parameter:

\begin{figure}[H]
\centering
\includegraphics[width=12 cm]{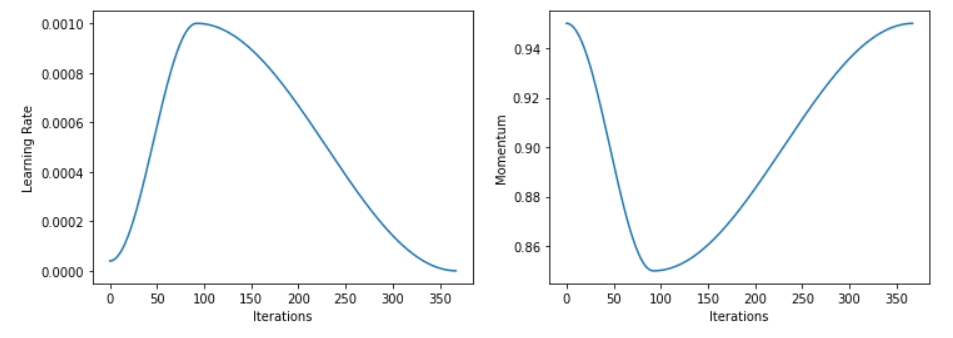}
\caption{The hyper-parameters schedule in the 1cycle policy.}
\end{figure}

The learning rate is the most important hyper-parameter to tune (and very often the only one since the library sets proper defaults). Other libraries often provide help for grid search or AutoML to guess the best value, but the fastai library implements the learning rate finder~\cite{lrfinder} which much more quickly provides the best value for this parameter after a mock training. The command \verb+learn.lr_find()+ will return a graph like this:

\begin{figure}[H]
\centering
\includegraphics[width=7 cm]{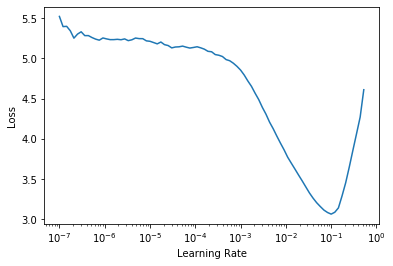}
\caption{The learning rate finder does a mock training with an exponentially growing learning rate over 100 iterations. A good value is then the minimum value on the graph divided by 10.}
\end{figure}

Another important high-level API component, which is shared across all of the applications, is the \emph{data block API}. The data block API is an expressive API for data loading. It is the first attempt we are aware of to systematically define all of the steps necessary to prepare data for a deep learning model, and give users a mix and match recipe book for combining these pieces (which we refer to as \emph{data blocks}). The steps that are defined by the data block API are: 

\begin{itemize}
    \item Getting the source items,
    \item Splitting the items into the training set and one or more validation sets,
    \item Labelling the items,
    \item Processing the items (such as normalization), and
    \item Optionally collating the items into batches.
\end{itemize}

Here is an example of how to use the data block API to get the MNIST dataset~\cite{mnist} ready for modelling:

\begin{lstlisting}[language=Python]
mnist = DataBlock(
    blocks=(ImageBlock(cls=PILImageBW), CategoryBlock), 
    get_items=get_image_files, 
    splitter=GrandparentSplitter(),
    get_y=parent_label)
dls = mnist.databunch(untar_data(URLs.MNIST_TINY), batch_tfms=Normalize)
\end{lstlisting}

In fastai v1 and earlier we used a fluent instead of a functional API for this (meaning the statements to execute those steps were chained one after the other). We discovered that this was a mistake; while fluent APIs are flexible in the order in which the user can define the steps, that order is very important in practice. With this functional \verb+DataBlock+ you don't have to remember if you need to split before or after labelling your data, for instance. Also, fluent APIs, at least in Python, tend not to work well with auto completion technologies. The data processing can be defined using Transforms (see \ref{transform}). Here is an example of using the data blocks API to complete the same segmentation seen earlier:

\begin{lstlisting}[language=Python]
path = untar_data(URLs.CAMVID_TINY)
camvid = DataBlock(blocks=(ImageBlock, ImageBlock(cls=PILMask)),
    get_items=get_image_files,
    splitter=RandomSplitter(),
    get_y=lambda o: path/'labels'/f'{o.stem}_P{o.suffix}')
dls = camvid.databunch(path/"images",
    batch_tfms=[*aug_transforms(), Normalize.from_stats(*imagenet_stats)])
\end{lstlisting}

Object detection can also be completed using the same functionality (here using the COCO dataset~\cite{coco}):

\begin{lstlisting}[language=Python]
coco_source = untar_data(URLs.COCO_TINY)
images, lbl_bbox = get_annotations(coco_source/'train.json')
lbl = dict(zip(images, lbl_bbox))

coco = DataBlock(blocks=(ImageBlock, BBoxBlock, BBoxLblBlock),
    get_items=get_image_files,
    splitter=RandomSplitter(),
    getters=[noop, lambda o:lbl[o.name][0], lambda o:lbl[o.name][1]], 
    n_inp=1)
dls = coco.databunch(coco_source, item_tfms=Resize(128),
    batch_tfms=[*aug_transforms(), Normalize.from_stats(*imagenet_stats)])
\end{lstlisting}

In this case, the targets are a tuple of two things: a list of bounding boxes and a list of labels. This is why there are three blocks, a list of getters and an extra argument to specify how many of the blocks should be considered the input (the rest forming the target).

The data for language modeling seen earlier can also be built using the data blocks API:

\begin{lstlisting}[language=Python]
df = pd.read_csv(untar_data(URLs.IMDB_SAMPLE)/'texts.csv')
df_tok,count = tokenize_df(df, 'text')
imdb_lm = DataBlock(blocks=TextBlock(make_vocab(count),is_lm=True),
                    get_x=attrgetter('text'),
                    splitter=RandomSplitter())
dls = imdb_lm.databunch(df_tok, bs=64, seq_len=72)
\end{lstlisting}

We have heard from users that they find the data blocks API provides a good balance of conciseness and expressivity. Many libraries have provided various approaches to data processing. In the data science domain the scikit-learn~\cite{scikit-learn} \emph{pipeline} approach is widely used. This API provides a very high level of expressivity, but is not opinionated enough to ensure that a user completes all of the steps necessary to get their data ready for modelling. As another example, TensorFlow~\cite{tensorflow} provides the tf.data library, which does not as precisely map the steps necessary for users to complete their task to the functionality provided by the API. The Torchvision~\cite{torchvision} library is a good example of an API which is highly specialised to a small subset of data processing tasks for a specific subdomain. fastai tries to capture the benefits of both extremes of the spectrum, without compromises; the data blocks API is how most users transform their data for use with the library.

\subsection{Incrementally adapting PyTorch code}\label{incremental}

Users often need to use existing pure PyTorch code (i.e. code that doesn't use fastai), such as their existing code-bases developed without fastai, or using third party code written in pure PyTorch. fastai supports incrementally adding fastai features to this code, without requiring extensive rewrites.

For instance, at the time of writing, the official PyTorch repository includes a MNIST training example\footnote{\url{https://github.com/pytorch/examples/blob/master/mnist/main.py}}. In order to train this example using fastai's \verb+Learner+ only two steps are required. First, the 30 lines in the example covering the \verb+test()+ and \verb+train()+ functions can be removed. Then, the 4 lines of the training loop is replaced with this code:

\begin{lstlisting}[language=Python]
data = DataLoaders(train_loader, test_loader).cuda()
learn = Learner(data, Net(), loss_func=F.nll_loss, opt_func=Adam, metrics=accuracy)
learn.fit_one_cycle(epochs, lr)
\end{lstlisting}

With no other changes, the user now has the benefit of all fastai's callbacks, progress reporting, integrated schedulers such as 1cycle training, and so forth.

\subsection{Consistency across domains}\label{consistency}

As the application examples have shown, the fastai library allows training a variety of kinds of application models, with a variety of kinds of datasets, using a very consistent API. The consistency covers not just the initial training, but also visualising and exploring the input data and model outputs. Such consistency helps students, both through having less to learn, and through showing the unifying concepts across different types of model. It also helps practitioners and researchers focus on their model development rather than learning incidental differences between APIs across domains. It is of particular benefit when, for instance, an NLP expert tries to bring their expertise across to a computer vision application.

There are many libraries that provide high-level APIs to specific applications, such as Facebook's Torchvision~\cite{torchvision}, Detectron~\cite{detectron}, and Fairseq~\cite{fairseq}. However, each library has a different API, input representation, and requires different assumptions about training details, all of which a user must learn from scratch each time. This means that there are many deep learning practitioners and researchers who become specialists in specific subfields, partially based on their understanding of the toiling of those subfields. By providing a consistent API fastai users are able to quickly move between different fields and reuse their expertise.

Customizing the behaviour of predefined applications can be challenging, which means that researchers often end up "reinventing the wheel", or, constraining themselves to the specific parts which there tooling allows them to customize. Because fastai provides a layered architecture, users of the software can customize every part, as they need. The layered architecture is also an important foundation in allowing PyTorch users to incrementally add fastai functionality to existing code bases. Furthermore, fastai's layers are reused across all applications, so an investment in learning them can be leveraged across many different projects.

The approach of creating layered APIs has a long history in software engineering. Software engineering best practices involve building up decoupled components which can be tied together in flexible ways, and then creating increasingly less abstract and more customized layers on top of each part.

The layered API design is also important for researchers and practitioners aiming to create best in class results. As the field of deep learning matures, there are more and more architectures, optimizers, data processing pipelines, and other approaches that can be selected from. Trying to bring multiple approaches together into a single project can be extremely challenging, when each one is using a different, incompatible API, and has different expectations about how a model is trained. For instance, in the original mixup article~\cite{mixup}, the code provided by the researchers only works on one specific dataset, with one specific set of metrics, and with one specific optimizer. Attempting to combine the researchers' mixup code with other training best practices, such as mixed precision training~\cite{mixedprecision}, requires rewriting it largely from scratch. The next section will look at the mid-level API pieces that fastai provides, which can be mixed and matched together to allow custom approaches to be quickly and reliably created.

\section{Mid-level APIs}

Many libraries, including fastai version 1 or earlier, provide a high-level API to users, and a low-level API used internally for that functionality, but nothing in between. This has two problems: the first is that it becomes harder and harder to create additional high-level functionality, as the system becomes more sophisticated, because the low-level API becomes increasingly complicated and cluttered. The second problem is that for users of the system who want to customize and adapt it, they often have to rewrite significant parts of the high-level API, and understand the large surface area of the low-level API in order to do so. This tends to mean that only a small dedicated community of specialists can really customize the software. 

These issues are common across nearly all software development, and many software engineers have worked hard to find ways to deal with this complexity and develop layered architectures. The issue in the deep learning community, however, is that these practices have not seemed to be widely understood or adopted. There are, however, exceptions; most relevant to this paper, the PyTorch library~\cite{pytorch} has a carefully layered design and is highly customizable. 

Much of the innovation in fastai is in its new mid-level APIs. This section will look at the following mid-level APIs: data, callbacks, optimizer, model layers, and metrics. These APIs are what the four fastai applications are built with and are also fully documented and available to users so that they can build their own applications or customize the existing ones. 

\subsection{Learner}\label{learner}

As already noted, a library can provide more appropriate defaults and user-friendly behaviour by ensuring that classes have all the information they need to make appropriate choices. One example of this is the \verb+DataLoaders+ class, which brings together all the information necessary for creating the data required for modelling. fastai also provides the \verb+Learner+ class, which brings together all the information necessary for training a model based on the data. The information which \verb+Learner+ requires, and is stored as state within a learner object, is: a PyTorch model, and optimizer, a loss function, and a \verb+DataLoaders+ object. Passing in the optimizer and loss function is optional, and in many situations fastai can automatically select appropriate defaults. 

\verb+Learner+ is also responsible (along with \verb+Optimizer+) for handling fastai's transfer learning functionality. When creating a \verb+Learner+ the user can pass a \emph{splitter}. This is a function that describes how to split the layers of a model into PyTorch parameter groups, which can then be frozen, trained with different learning rates, or more generally handled differently by an optimizer. 

One area that we have found particularly sensitive in transfer learning is the handling of batch-normalization layers~\cite{batchnorm}. We tried a wide variety of approaches to training and updating the moving average statistics of those layers, and different configurations could often change the error rate by as much as 300\%. There was only one approach that consistently worked well across all datasets that we tried, which is to never freeze batch-normalization layers, and never turn off the updating of their moving average statistics. Therefore, by default, \verb+Learner+ will bypass batch-normalization layers when a user asks to freeze some parameter groups. Users often report that this one minor tweak dramatically improves their model accuracy and is not something that is found in any other libraries that we are aware of. 

\verb+DataLoaders+ and \verb+Learner+ also work together to ensure that model weights and input data are all on the same device. This makes working with GPUs significantly more straightforward and makes it easy to switch from CPU to GPU as needed. 

\subsection{Two-way callbacks}\label{callback}

In fastai version 0.7, we repeatedly modified the training loop in Learner to support many different tweaks and customizations. Over time, however, this became unwieldy. We noticed that there was a core subset of functionality that appeared in every one of these tweaks, and that all the other changes that were required could be refactored into a specific set of customization points. In other words, a wide variety of training methods can be represented using a single, universal training system. Once we extracted those common pieces, we were left with the basic fastai training loop, and the customisation points that we call \emph{two-way callbacks}. 

The \verb+Learner+ class's novel 2-way callback system allows gradients, data, losses, control flow, and anything else to be read and changed at any point during training. There is a rich history of using callbacks to allow for customisation of numeric software, and today nearly all modern deep learning libraries provide this functionality. However, fastai's callback system is the first that we are aware of that supports the design principles necessary for complete two-way callbacks:

\begin{itemize}
    \item A callback should be available at every single point that code can be run during training, so that a user can customise every single detail of the training method ;
    \item Every callback should be able to access every piece of information available at that stage in the training loop, including hyper-parameters, losses, gradients, input and target data, and so forth ;
\end{itemize}

This is the way callbacks are usually designed, but in addition, there is a key design principle: 

\begin{itemize}
    \item Every callback should be able to modify all these pieces of information, at any time before they are used, and be able to skip a batch, epoch, training or validation section, or cancel the whole training loop.
\end{itemize}

This is why we call these 2-way callbacks, as the information not only flows from the training loop to the callbacks, but on the other way as well. For instance, here is the code for training a single batch \verb+b+ in fastai: 

\begin{lstlisting}[language=Python]
try:
  self._split(b);                                 self.cb('begin_batch')
  self.pred = self.model(*self.x);                self.cb('after_pred')
  if len(self.y) == 0: return
  self.loss = self.loss_func(self.pred, *self.y); self.cb('after_loss')
  if not self.training: return
  self.loss.backward();                           self.cb('after_back')
  self.opt.step();                                self.cb('after_step')
  self.opt.zero_grad()
except CancelBatchException:                      self.cb('after_cancel')
finally:                                          self.cb('after_batch')
\end{lstlisting}

This example clearly shows how every step of the process is associated with a callback (the calls to \verb+self.cb()+ and shows how exceptions are used as a flexible control flow mechanism for them. 

In fastai v0.7, we did not follow these three design principles. As a result, we had to frequently change the training loop to support additional functionality, and new research papers. On the other hand, with this new callback system we have not had to change the training loop at all, and have used callbacks to implement mixup augmentation, generative adversarial networks, optimized mixed precision training, PyTorch hooks, the learning rate finder, and many more. Most importantly, we have not yet come across any cases where mixing and matching these callbacks has caused any problems. Therefore, users can use all the training features that they want, and can easily do ablation studies, adding, removing, and modifying techniques as needed. 

\subsubsection{Case study: generative adversarial network training using callbacks}

A good example of a fastai callback is \verb+GANTrainer+, which implements training of generative adversarial networks~\cite{NIPS2014_5423}. To do so, it must complete the following tasks:

\begin{itemize}
    \item Freeze the generator and train the critic for one (or more) step by: 
    \begin{itemize}
        \item getting one batch of "real" images ; 
        \item generating one batch of "fake" images; 
        \item have the critic evaluate each batch and compute a loss function from that, which rewards positively the detection of real images and penalizes the fake ones; 
        \item update the weights of the critic with the gradients of this loss.
    \end{itemize}
    \item Freeze the critic and train the generator for one (or more) step by: 
    \begin{itemize}
        \item generating one batch of "fake" images; 
        \item evaluate the critic on it; 
        \item return a loss that rewards positively the critic thinking those are real images;
        \item update the weights of the generator with the gradients of this loss.
    \end{itemize}
\end{itemize}

To do so, it relies on a \verb+GANModule+  that contains the generator and the critic, then delegates the input to the proper model depending on the value of a flag \verb+gen_mode+ and on a \verb+GANLoss+ that also has a generator or critic behavior and handles the evaluation mentioned earlier. Then, it defines the following callback methods: 

\begin{itemize}
    \item \verb+begin_fit+: Initialises the generator, critic, loss functions, and internal storage 
    \item \verb+begin_epoch+: Sets the critic or generator to training mode
    \item \verb+begin_validate+: Switches to generator mode for showing results
    \item \verb+begin_batch+: Sets the appropriate target depending on whether it is in generator or critic mode 
    \item \verb+after_batch+: Records losses to the generator or critic log 
    \item \verb+after_epoch+: Optionally shows a sample image 
\end{itemize}

This callback is then customized with another callback, which defines at what point to switch from critic to generator and vice versa. fastai includes several possibilities for this purpose, such as an \verb+AdaptiveGANSwitcher+, which automatically switches between generator and critic training based on reaching certain thresholds in their respective losses. This approach to training can allow models to be trained significantly faster and more easily than with standard fixed schedule approaches. 

\subsection{Generic optimizer}\label{optimizer}

fastai provides a new generic optimizer foundation that allows recent optimization techniques to be implemented in a handful of lines of code, by refactoring out the common functionality of modern optimizers into two basic pieces: 

\begin{itemize}
    \item \emph{stats}, which track and aggregate statistics such as gradient moving averages ;
    \item \emph{steppers}, which combine stats and hyper-parameters to update the weights using some function.
\end{itemize}

This has allowed us to implement every optimizer that we have attempted in fastai, without needing to extend or change this foundation. This has been very beneficial, both for research and development. As an example of a development improvement, here are the entire changes needed to make to support decoupled weight decay (also known as AdamW~\cite{adamw}): 

\begin{lstlisting}[language=Python]
steppers = [weight_decay] if decouple_wd else [l2_reg]
\end{lstlisting}

On the other hand, the implementation in the PyTorch library required creating an entirely new class, with over 50 lines of code. The benefit for research comes about because it it easy to rapidly implement new papers as they come out, recognise similarities and differences across techniques, and try out variants and combinations of these underlying differences, many of which have not yet been published. The resulting code tends to look a lot like the maths shown in the paper. For instance, here is the code in fastai, and the algorithm from the paper, for the LAMB optimizer~\cite{lamb}: 

\begin{figure}[H]
\centering
\includegraphics[width=15 cm]{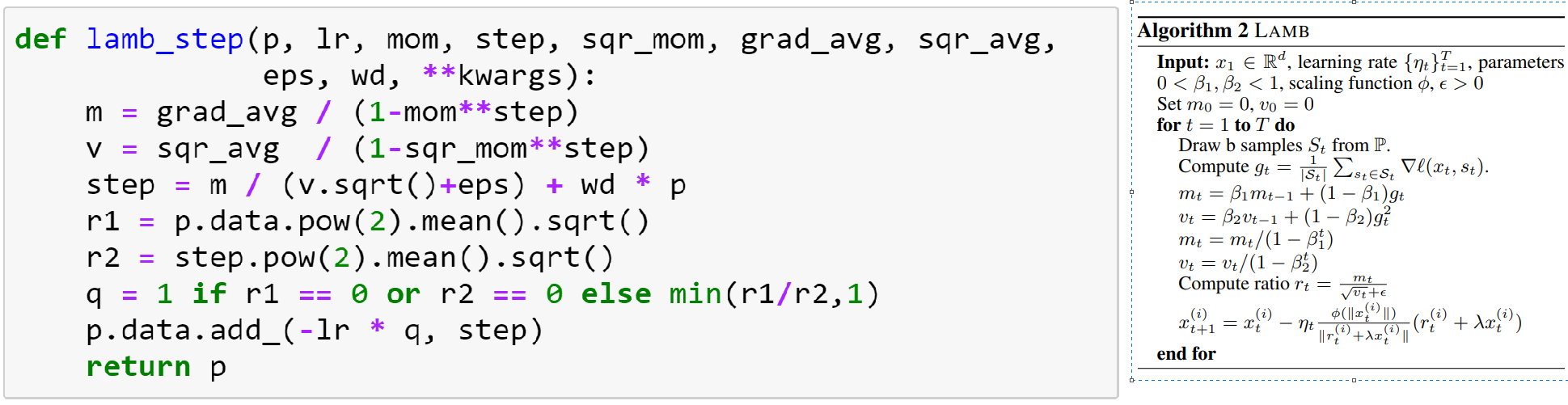}
\caption{The LAMB algorithm and implementation.}
\end{figure} 

The only difference between the code and the figure are:
\begin{itemize}
    \item the means that update $m_{t}$ and $v_{t}$ don't appear as this done in a separate function \emph{stat};
    \item the authors do not provide the full definition of the $\phi$ function they use (it depends on undefined parameters), the code below is based on the official TensorFlow implementation.
\end{itemize}

In order to support modern optimizers such as LARS fastai allows the user to choose whether to aggregate stats at model, layer, or per activation level. 

\subsection{Generalized metric API}\label{metric}

Nearly all machine learning and deep learning libraries provide some support for \emph{metrics}. These are generally defined as simple functions which take the mean, or in some cases a custom reduction function, across some measurement which is logged during training. However, some metrics cannot be correctly defined using this framework. For instance, the dice coefficient, which is widely used for measuring segmentation accuracy, cannot be directly expressed using a simple reduction. 

In order to provide a more flexible foundation to support metrics like this fastai provides a \verb+Metric+ abstract class which defines three methods: \verb+reset+, \verb+accumulate+, and \verb+value+ (which is a property). Reset is called at the start of training, accumulate is called after each batch, and then finally value is called to calculate the final check. Whenever possible, we can thus avoid recording and storing all predictions in memory. For instance, here is the definition of the dice coefficient: 

\begin{lstlisting}[language=Python]
class Dice(Metric):
    def __init__(self, axis=1): self.axis = axis
    def reset(self): self.inter,self.union = 0,0
    def accumulate(self, learn):
        pred,targ = flatten_check(learn.pred.argmax(self.axis), learn.y)
        self.inter += (pred*targ).float().sum().item()
        self.union += (pred+targ).float().sum().item()

    @property
    def value(self): return 2. * self.inter/self.union if self.union>0 else None
\end{lstlisting}

The Scikit-learn library~\cite{scikit-learn} already provides a wide variety of useful metrics, so instead of reinventing them, fastai provides a simple wrapper function, \verb+skm_to_fastai+, which allows them to be used in fastai, and can automatically add pre-processing steps such as sigmoid, argmax, and thresholding. 

\subsection{fastai.data.external}

Many libraries have recently started integrating access to external datasets directly into their APIs. fastai builds on this trend, by curating and collecting a number of datasets (hosted by the AWS Public Dataset Program\footnote{\url{https://aws.amazon.com/opendata/public-datasets/}}) in a single place and making them available through the \verb+fastai.data.external+ module. fastai automatically downloads, extracts, and caches these datasets when they are first used. This is similar to functionality provided by Torchvision, TensorFlow datasets, and similar libraries, with the addition of closer integration into the fastai ecosystem. For instance, fastai provides cut-down “sample” versions of many of its datasets, which are small enough that they can be downloaded and used directly in documentation, continuous integration testing, and so forth. These datasets are also used in the documentation, along with examples showing users what they can expect when training models with his datasets. Because the documentation is written in interactive notebooks (as discussed in a later section) this also means that users can directly experiment with these datasets and models by simply running and modifying the documentation notebooks. 

\subsection{funcs\_kwargs and DataLoader}\label{dataloader}

Once a user has their data available, they need to get it into a form that can be fed to a PyTorch model. The most common class used to feed models directly is the \verb+DataLoader+ class in PyTorch. This class provides fast and reliable multi-threaded data-processing execution, with several points allowing customisation. However, we have found that it is not flexible enough to conveniently do some of the tasks that we have required, such as building a \verb+DataLoader+ for an NLP language model. Therefore, fastai includes a new \verb+DataLoader+ class on top of the internal classes that PyTorch uses. This combines the benefits of the fast and reliable infrastructure provided by PyTorch with a more flexible and expressive front-end for the user. 

\verb+DataLoader+ provides 15 extension points via customizable methods, which can be replaced by the user as required. These customizable methods represent the 15 stages of data loading that we have identified, and which fit into three broad stages: sample creation, item creation, and batch creation. In contrast, in the standard PyTorch \verb+DataLoader+ class only a small subset of these stages is explicitly made available for customization by the user. Unless a user's requirements are met by this subset, the user is forced to implement their own solution from scratch. The impact of this additional customizability can be quite significant. For instance, the fastai language model \verb+DataLoader+ went from 90 lines of code to 30 lines of code after adopting this approach. 

What makes this flexibility possible is a Python decorator that is called \verb+funcs_kwargs+. This decorator creates a class in which any method can be replaced by passing a new function to the constructor, or by replacing it through subclassing. This allows users to replace any part of the logic in the \verb+DataLoader+ class. In order to maximise the power of this, nearly every part of the fastai \verb+DataLoader+ is a method with a single line of code. Therefore, virtually every design choice can be adjusted by users. 

fastai also provides a transformed \verb+DataLoader+ called \verb+TfmdDL+, which subclasses \verb+DataLoader+. In \verb+TfmdDL+ the callbacks and customization points execute \emph{Pipelines} of \emph{Transforms}. Both mechanisms are described in \ref{transform}; this section provides a brief overview here. A \verb+Transform+ is simply a Python function, which can also include its inverse function -- that is, the function which “undoes” the transform. Transforms can be composed using the \verb+Pipeline+ class, which then allows the entire function composition to be inverted as well. We refer to these two directions, the forward and inverse directions of the functions, as the \verb+Transform+s' \verb+encodes+ and \verb+decodes+ methods.

\verb+TfmdDL+ provides the foundations for the visualisation support discussed in the application section, having the basic template for showing a batch of data. In order to do this, it needs to \emph{decode} any transforms in the pipeline, which it does automatically. For instance, an integer representing a level of a category will be converted back into the string that the integer represents. 

\subsection{fastai.data.core}\label{datacore}

When users who need to create a new kind of block for the data blocks API, or need a level of customization that even the data blocks API doesn't support, they can use the mid-level components that the data block API is built on. These are a small number of simple classes which combine the transform pipelines functionality of fastai with Python's collections interface. 

The most basic class is transformed list, or \verb+TfmdLists+, which lazily applies a transform pipeline to a collection, whilst providing a standard Python collection interface. This is an important foundational functionality for deep learning, such as the ability to index into a collection of filenames, and on demand read an image file then apply any processing, such as data augmentation and normalization, necessary for a model. \verb+TfmdLists+ also provides \emph{subset} functionality, which allows the user to define subsets of the data, such as those representing training and validation sets. Information about what subset an item belongs to is passed down to transforms, so that they can ensure that they do the appropriate processing -- for instance, data augmentation processing would be generally skipped for a validation set, unless doing test time augmentation.

Another important data class at this layer of the API is \verb+Datasets+, which applies multiple transform pipelines in parallel to a single collection. Like \verb+TfmdLists+, it provides a standard Python collection interface. Indexing into a \verb+Datasets+ object returns a couple containing the result of each transform pipeline on the input item. This is the class used by the data blocks API to return, for instance, a tuple of an image tensor, and a label for that image, both derived from the same input filename. 

\subsection{Layers and architectures}\label{layer}

PyTorch (like many other libraries) provides a basic “sequential” layer object, which can be combined in sequence to form a component of a network. This represents simple composition of functions, where each layer's output is the next layer's input. However, many components in modern network architectures cannot be represented in this way. For instance, ResNet blocks~\cite{resnet}, and any other block which requires a skip connection, are not compatible with sequential layers. The normal workaround for this in PyTorch is to write a custom forward function, effectively relying on the full flexibility of Python to escape the limits of composing these sequence layers. 

However, there is a significant downside: the model is now no longer amenable to easy analysis and modification, such as removing the final few layers in order to do transfer learning. This also makes it harder to support automatic drawing graphs representing a model, printing a model summary, accurately reporting on model computation requirements by layer, and so forth. 

Therefore, fastai attempts to provide the basic foundations to allow modern neural network architectures to be built by stacking a small number of predefined building blocks. The first piece of this system is the \verb+SequentialEx+ layer. This layer has the same basic API as PyTorch's \verb+nn.Sequential+, with one key difference: the original input value to the function is available to every layer in the block. Therefore, the user can, for instance, include a layer which adds the current value of the sequential block to the input value of the sequential block (such as is done in a ResNet). 

To take full advantage of this capability, fastai also provides a \verb+MergeLayer+ class. This allows the user to pass any function, which will in turn be provided with the layer block input value, and the current value of the sequential block. For instance, if you pass in a simple add function, then \verb+MergeLayer+ provides the functionality of an identity connection in a standard resnet block. Or, if the user passes in a concatenation function, then it provides the basic functionality of a concatenating connection in a Densenet block~\cite{densenet}. In this way, fastai provides primitives which allow representing modern network architecture out of predefined building blocks, without falling back to Python code in the forward function. 

fastai also provides a general-purpose class for combining these layers into a wide range of modern convolutional neural network architectures. These are largely based on the underlying foundations from ResNet~\cite{resnet}, and therefore this class is called \verb+XResNet+. By providing parameters to this class, the user can customise it to create architectures that include squeeze and excitation blocks~\cite{senet}, grouped convolutions such as in ResNext~\cite{resnext}, depth-wise convolutions such as in the Xception architecture~\cite{xception}, widening factors such as in Wide ResNets~\cite{wideresnet}, self-attention and symmetric self-attention functionality , custom activation functions, and more. By using this generic refactoring of these clusters of modern neural network architectures, we have been able to design and experiment with novel combinations very easily. It is also clearer to users exactly what is going on in their models, because the various specific architectures are clearly represented as changes to input parameters. 

One set of techniques that is extremely useful in practice are the tweaks to the ResNet architecture described in~\cite{bagoftricks}. These approaches are used by default in \verb+XResNet+. Another architecture tweak which has worked well in many situations is the recently developed Mish activation function~\cite{mish}. fastai includes an implementation of Mish which is optimised using PyTorch's just-in-time compiler (JIT). 

A similar approach has been used to refactor the U-Net architecture~\cite{unet}. Through looking at a range of competition winning and state-of-the-art papers in segmentation, we curated a set of approaches that work well together in practice. These are made available by default in fastai's U-Net implementation, which also dynamically creates the U-Net cross connections for any given input size. 

\section{Low-level APIs}

The layered approach of the fastai library has a specific meaning at the lower levels of it stack. Rather than treating Python~\cite{python37} itself as the base layer of the computation, which the middle layer relies on, those layers rely on a set of basic abstractions provided by the lower layer. The middle layer is programmed in that set of abstractions. The low-level of the fastai stack provides a set of abstractions for: 

\begin{itemize}
    \item Pipelines of transforms: Partially reversible composed functions mapped and dispatched over elements of tuples
    \item Type-dispatch based on the needs of data processing pipelines
    \item Attaching semantics to tensor objects, and ensuring that these semantics are maintained throughout a \verb+Pipeline+
    \item GPU-optimized computer vision operations
    \item Convenience functionality, such as a decorator to make patching existing objects easier, and a general collection class with a NumPy-like API.
\end{itemize}

The rest of this section will explain how the transform pipeline system is built on top of the foundations provided by PyTorch, type dispatch, and semantic tensors, providing the flexible infrastructure needed for the rest of fastai.

\subsection{PyTorch foundations}

The main foundation for fastai is the PyTorch~\cite{pytorch} library. PyTorch provides a GPU optimised tensor class, a library of useful model layers, classes for optimizing models, and a flexible programming model which integrates these elements. fastai uses building blocks from all parts of the PyTorch library, including directly patching its tensor class, entirely replacing its library of optimizers, providing simplified mechanisms for using its hooks, and so forth. In earlier prototypes of fastai we used TensorFlow~\cite{tensorflow} as our platform (and before that used~\cite{theano}), but switched to PyTorch because we found that it had a fast core, a simple and well curated API, and rapidly growing popularity in the research community. At this point most papers at the top deep learning conferences are implemented using PyTorch.

fastai builds on many other open source libraries. For CPU image processing fastai uses and extends the Python imaging library (PIL)~\cite{pil}, for reading and processing tabular data it uses pandas, for most of its metrics it uses Scikit-Learn~\cite{scikit-learn}, and for plotting it uses Matplotlib~\cite{matplotlib}. These are the most widely used libraries in the Python open source data science community and provide the features necessary for the fastai library.

\subsection{Transforms and Pipelines}\label{transform}

One key motivation is the need to often be able to undo some subset of transformations that are applied to create the data used to modelling. This strings that represent categories cannot be used in models directly and are turned into integers using some vocabulary. And pixel values for images are generally normalized. Neither of these can be directly visualized, and therefore at inference time we need to apply the inverse of these functions to get data that is understandable. Therefore, fastai introduces a \verb+Transform+ class, which provides callable objects, along with a \emph{decode} method. The decode method is designed to invert the function provided by a transform; it needs to be implemented manually by the user ; it is similar to the \verb+inverse_transform+ you can provide in Scikit-Learn~\cite{scikit-learn} pipelines and transformers. By providing both the encode and decode methods in a single place, the user ends up with a single object which they can compose into pipelines, serialize, and so forth. 

Another motivation for this part of the API is the insight that PyTorch data loaders provide tuples, and PyTorch models expect tuples as inputs. Sometimes these tuples should behave in a connected and dependent way, such as in a segmentation model, where data augmentation must be applied to both the independent and dependent variables in the same basic way. Sometimes, however, different implementations must be used for different types; for instance, when doing affine transformations to a segmentation mask nearest-neighbor interpolation is needed, but for an image generally a smoother interpolation function would be used.

In addition, sometimes transforms need to be able to opt out of processing altogether, depending on context. For instance, except when doing test time augmentation, data augmentation methods should not be applied to the validation set. Therefore, fastai automatically passes the current subset index to transforms, allowing them to modify their behaviour based on subset (for instance, training versus validation). This is largely hidden from the user, because base classes are provided which automatically do this context-dependent skipping. However, advanced users requiring complete customization can use this functionality directly. 

Transforms in deep learning pipelines often require state, which can be dependent on the input data. For example, normalization statistics could be based on a sample of data batches, a categorization transform could get its vocabulary directly from the dependent variable, or an NLP numericalization transform could get its vocabulary from the tokens used in the input corpus. Therefore, fastai transforms and pipelines support a \emph{setup} method, which can be used to create this state when setting up a \verb+Pipeline+. When pipelines are set up, all previous transforms in the pipeline are run first, so that the transform being set up receives the same structure of data that it will when being called.  

This is closely connected to the implementation of \verb+TfmdList+. Because a \verb+TfmdList+ lazily applies a pipeline to a collection, fastai can automatically call the \verb+Pipeline+ \verb+setup+ method as soon as it is connected to the collection in a \verb+TfmdList+. 

\subsection{Type dispatch}\label{typedispatch}

The fastai type dispatch system is like the \verb+functools.singledispatch+ system provided in the Python standard library while supporting multiple dispatch over two parameters. Dispatch over two parameters is necessary for any situation where the user wants to be able to customize behavior based on both the input and target of a model. For instance, fastai uses this for the \verb+show_batch+ and \verb+show_results+ methods. As shown in the application section, these methods automatically provide an appropriate visualisation of the input, target, and results of a model, which requires responding to the types of both parameters. In one example the input was an image, and the target was a segmentation mask, and the show results method automatically used a colour-coded overlay for the mask. On the other hand, for an image classification problem, the input would be shown as an image, the prediction and target would be shown as text labels, and color-coded based on whether they were correct. 

It also provides a more expressive and yet concise syntax for registering additional dispatched functions or methods, taking advantage of Python's recently introduced type annotations syntax. Here is an example of creating two different methods which dispatch based on parameter types: 

\begin{lstlisting}[language=Python]
@typedispatch
def f_td_test(x:numbers.Integral, y): return x+1
@typedispatch
def f_td_test(x:int, y:float): return x+y
\end{lstlisting}

Here \verb+f_td_test+ has a generic implementation for \verb+x+ of numeric types and all \verb+y+s, then a specialized implementation when \verb+x+ is an \verb+int+ and \verb+y+ is a \verb+float+.

\subsection{Object-oriented semantic tensors}\label{ootensor}

By using fastai's transform pipeline functionality, which depends heavily on types, the mid and high-level APIs can provide a lot of power, conciseness, and expressivity for users. However, this does not work well with the types provided by PyTorch, since the basic tensor type does not have any subclasses which can be used for type dispatch. Furthermore, subclassing PyTorch tensors is challenging, because the basic functionality for instantiating the subclasses is not provided and doing any kind of tensor operation will strip away the subclass information. 

Therefore, fastai provides a new tensor base class, which can be easily instantiated and subclass. fastai also patches PyTorch's tensor class to attempt to maintain subclass information through operations wherever possible. Unfortunately, it is not possible to always perfectly maintain this information throughout every possible operation, and therefore all fastai \verb+Transform+ automatically maintain subclass types appropriately. 

fastai also provides the same functionality for Python imaging library classes, along with some basic type hierarchies for Python built-in collection types, NumPy arrays, and so forth. 

\subsection{GPU-accelerated augmentation}\label{gpuaug}

The fastai library provides most data augmentation in computer vision on the GPU at the batch level. Historically, the processing pipeline in computer vision has always been to open the images and apply data augmentation on the CPU, using a dedicated library such as PIL~\cite{pil} or OpenCV~\cite{opencv}, then batch the results before transferring them to the GPU and using them to train the model. On modern GPUs however, architectures like a standard ResNet-50 are often CPU-bound. Therefore fastai implements most common functions on the GPU, using PyTorch's implementation of \verb+grid_sample+ (which does the interpolation from the coordinate map and the original image).

Most data augmentations are random affine transforms (rotation, zoom, translation, etc), functions on a coordinates map (perspective warping) or easy functions applied to the pixels (contrast or brightness changes), all of which can easily be parallelized and applied to a batch of images. In fastai, we combine all affine and coordinate transforms in one step to only apply one interpolation, which results in a smoother result. Most other vision libraries do not do this and lose a lot of detail of the original image when applying several transformations in a row. 
 
\begin{figure}[H]
\centering
\includegraphics[width=12 cm]{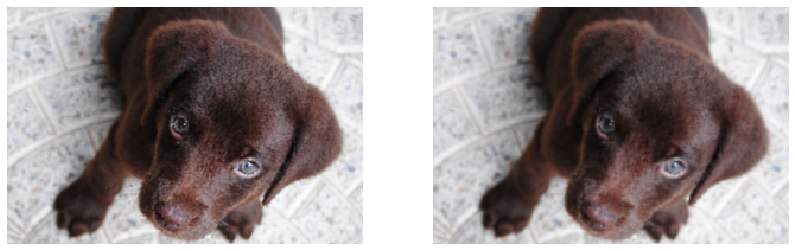}
\caption{A rotation and a zoom apply to an image with one interpolation only (right) or two interpolations (left). The latter results in more texture loss.}
\end{figure} 

The type-dispatch system helps apply appropriate transforms to images, segmentation masks, key-points or bounding boxes (and users can add support for other types by writing their own functions).

\subsection{Convenience functionality}

fastai has a few more additions designed to make Python easier to use, including a NumPy-like API for lists called \verb+L+, and some decorators to make delegation or patching easier.

Delegation is used when one function will call another and send it a bunch of keyword arguments with defaults. To avoid repeating those, they are often grouped into \verb+**kwargs+. The problem is that they then disappear from the signature of the function that delegates, and you can't use the tools from modern IDEs to get tab-completion for those delegated arguments or see them in its signature. To solve this, fastai provides a decorator called \verb+@delegates+ that will analyze the signature of the delegated function to change the signature of the original function. For instance the initialization of \verb+Learner+ has 11 keyword-arguments, so any function that creates a \verb+Learner+ uses this decorator to avoid mentioning them all. As an example, the function \verb+tabular_learner+ is defined like this:

\begin{lstlisting}[language=Python]
@delegates(Learner.__init__)
def tabular_learner(dls, layers, emb_szs=None, config=None, **kwargs):
\end{lstlisting}

but when you look at its signature, you will see the 11 additional arguments of \verb+Learner.__init__+ with their defaults.

Monkey-patching is an important functionality of the Python language when you want to add functionality to existing objects. fastai makes it easier and more concise with a \verb+@patch+ decorator, using Python's type-annotation system. For instance, here is how fastai adds the \verb+write()+ method to the \verb+pathlib.Path+ class:

\begin{lstlisting}[language=Python]
@patch
def write(self:Path, txt, encoding='utf8'):
    self.parent.mkdir(parents=True,exist_ok=True)
    with self.open('w', encoding=encoding) as f: f.write(txt)
\end{lstlisting}

Lastly, inspired by the NumPy~\cite{numpy} library, fastai provides a collection type, called \verb+L+, that supports fancy indexing and has a lot of methods that allow users to write simple expressive code. For example, the code below takes a list of pairs, selects the second item of each pair, takes its absolute value, filters items greater than 4, and adds them up:

\begin{lstlisting}[language=Python]
d = dict(a=1,b=-5,d=6,e=9).items()
L(d).itemgot(1).map(abs).filter(gt(4)).sum()
\end{lstlisting}

\verb+L+ uses context-dependent functionality to simplify user code. For instance, the \verb+sorted+ method can take any of the following as a key: a callable (sorts based on the value of calling the key with the item), a string (used as an attribute name), or an int (used as an index).

\section{nbdev}\label{nbdev}

\begin{figure}[H]
\centering
\fbox{\includegraphics[width=12 cm]{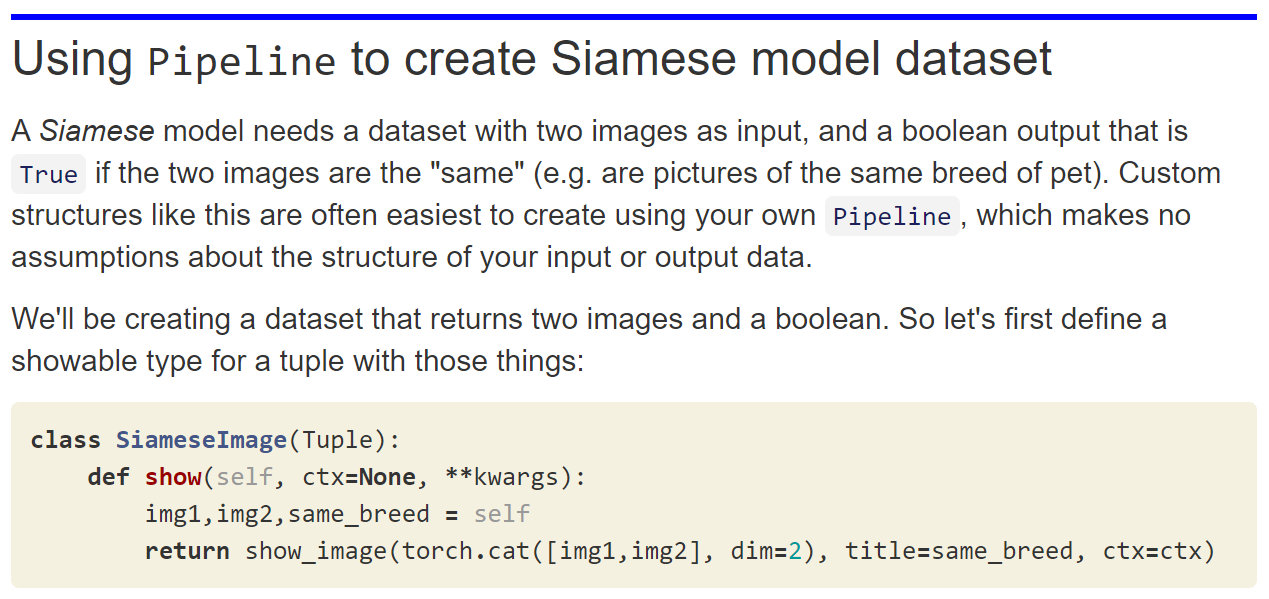}}
\caption{Example of fastai's documentation, automatically generated using nbdev}
\label{fig:nbdev}
\end{figure}

In order to assist in developing this library, we built a programming environment called nbdev, which allows users to create complete Python packages, including tests and a rich documentation system, all in Jupyter Notebooks~\cite{jupyter}. nbdev is a system for \textit{exploratory programming}. Exploratory programming is based on the observation that most developers spend most of their time as coders exploring and experimenting. Exploration is easiest developing on the prompt (or REPL), or using a notebook-oriented development system like Jupyter Notebooks. But these systems are not as strong for the “programming” part, since they're missing features provided by IDEs and editors like good documentation lookup, good syntax highlighting, integration with unit tests, and (most importantly) the ability to produce final, distributable source code files.

nbdev is built on top of Jupyter Notebook and adds the following critically important tools for software development:
\begin{itemize}
\item Python modules are automatically created, following best practices such as automatically defining \verb+__all__+ with exported functions, classes, and variables
\item Navigate and edit code in a standard text editor or IDE, and export any changes automatically back into your notebooks
\item Automatically create searchable, hyperlinked documentation from your code (as seen in figure ~\ref{fig:nbdev}; any word surrounded in backticks will by hyperlinked to the appropriate documentation, a sidebar is created in the documentation site with links to each of module, and more
\item Pip installers (uploaded to pypi automatically)
\item Testing (defined directly in notebooks, and run in parallel)
\item Continuous integration
\item Version control conflict handling
\end{itemize}

We plan to provide more information about the features, benefits, and history behind nbdev in a future paper.

\section{Related work}

There has been a long history of high-level APIs for deep learning in Python, and this history has been a significant influence on the development of fastai. The first example of a Python library for deep learning that we have found is Calysto/conx, which implemented back propagation in Python in 2001. Since that time there have been dozens of approaches to high-level APIs with perhaps the most significant, in chronological order, being Lasagne~\cite{lasagne} (begun in 2013), Fuel/Blocks (begun in 2014), and Keras~\cite{keras} (begun in 2015). There have been other directions as well, such as the configuration-based approach popularized by Caffe~\cite{caffe}, and lower-level libraries such as Theano~\cite{theano}, TensorFlow~\cite{tensorflow} and PyTorch~\cite{pytorch}. 

APIs from general machine learning libraries have also been an important influence on fastai. SPSS and SAS provided many utilities for data processing and analysis since the early days of statistical computing. The development of the S language was a very significant advance, which led directly  to projects such as R~\cite{rlanguage}, SPLUS, and xlisp-stat~\cite{xlispstat}. The direction taken by R for both data processing (largely focused on the “Tidyverse”~\cite{tidyverse}) and model building (built on top of R's rich “formula” system) shows how a very different set of design choices can result in a very different (and effective) user experience. Scikit-learn~\cite{scikit-learn}, Torchvision~\cite{torchvision}, and pandas~\cite{pandas} are examples of libraries which provide a function composition abstraction that (like fastai's \verb+Pipeline+) are designed to help users process their data into the format they need (Scikit-Learn also being able to perform learning and predictions on that processed data). There are also projects such as MLxtend~\cite{mlxtend} that provide a variety of utilities building on the functionality of their underlying programming languages (Python, in the case of MLxtend). 

The most important influence on fastai is, of course, PyTorch~\cite{pytorch}; fastai would not have been possible without it. The PyTorch API is extensible and flexible, and the implementation is efficient. fastai makes heavy use of \verb+torch.Tensor+ and \verb+torch.nn+ (including \verb+torch.nn.functional+). On the other hand, fastai does not make much use of PyTorch's higher level APIs, such as \verb+nn.optim+ and annealing, instead independently creating overlapping functionality based on the design approaches and goals described above. 

\section{Results and conclusion}

Early results from using fastai are very positive. We have used the fastai library to rewrite the entire fast.ai course “Practical Deep Learning for Coders”, which contains 14 hours of material, across seven modules, and covers all the applications described in this paper (and some more). We found that we were able to replicate or improve on all the results in previous versions of the material and were able to create the data pipelines and models needed much more quickly and easily than we could before. We have also heard from early adopters of pre-release versions of the library that they have been able to more quickly and easily write deep learning code and build models than with previous versions. fastai has already been selected as part of the official PyTorch Ecosystem\footnote{\url{https://pytorch.org/ecosystem/}}. According to the 2019 Kaggle ML \& DS Survey\footnote{\url{https://www.kaggle.com/c/kaggle-survey-2019}}, 10\% of data scientists in the Kaggle community are already using fastai. Many researchers are using fastai to support their work (e.g.~\cite{1905.04348}~\cite{1810.09025}~\cite{1906.11282}~\cite{1904.09076}).

Based on our experience with fastai, we believe that using a layered API in deep learning has very significant benefits for researchers, practitioners, and students. Researchers can see links across different areas more easily, rapidly combine and restructure ideas, and run experiments on top of strong baselines. Practitioners can quickly build prototypes, and then build on and optimize those prototypes by leveraging fastai's PyTorch foundations, without rewriting code. Students can experiment with models and try out variations, without being overwhelmed by boilerplate code when first learning ideas.

The basic ideas expressed in fastai are not limited to use in PyTorch, or even Python. There is already a partial port of fastai to Swift, called SwiftAI~\cite{swiftai}, and we hope to see similar projects for more languages and libraries in the future.

\authorcontributions{Both authors contributed equally to this work.}

\funding{This research received no external funding. Sylvain has been sponsored by AWS in 2018-2019 and by Google in 2019-2020.}

\acknowledgments{We would like to express our deep appreciation to Alexis Gallagher, who was instrumental throughout the paper-writing process, and who inspired the functional-style data blocks API. Many thanks also to the Facebook PyTorch team for all their support throughout fastai's development, to the global fast.ai community who through forums.fast.ai have contributed many ideas and pull requests that have been invaluable to the development of fastai, to Chris Lattner and the Swift for TensorFlow team who helped develop the Swift courses at course.fast.ai and SwiftAI, to Andrew Shaw for contributing to early prototypes of showdoc in nbdev, to Stas Bekman for contributing to early prototypes of the git hooks in nbdev and to packaging and utilities, and to the developers of the Python programming language, which provides such a strong foundation for fastai's features.}

\conflictsofinterest{The authors declare no conflict of interest.} 

\reftitle{References}
\externalbibliography{yes}
\bibliography{bibliography}


\end{document}